\documentclass{article}

\usepackage{amsmath}
\usepackage{amssymb}
\usepackage{amsfonts}
\usepackage{microtype}
\usepackage{graphicx}
\usepackage{sidecap}
\usepackage{subfigure}
\usepackage{booktabs} 
\usepackage{amsthm} 
\RequirePackage[utf8]{inputenc} 
\usepackage[pdfa]{hyperref} 
\newcommand{\LL}{\mathcal{L}}
\newcommand{\gd}[3] {\mathcal{N}_{#1}\left(#2, #3\right)}
\newcommand{\gdsmall}[3] {\mathcal{N}_{#1}(#2, #3)}
\newcommand{\argmin}{\mathop{\mathrm{argmin}}}
\newtheorem{proposition}{Proposition}

\usepackage[accepted]{icml2019}
\renewcommand{\algorithmiccomment}[1]{\bgroup\hfill\footnotesize~#1\egroup} 

\icmltitlerunning{APT for Likelihood-free Inference}

\begin{document}

\twocolumn[
\icmltitle{Automatic Posterior Transformation for Likelihood-free Inference}

\begin{icmlauthorlist}
\icmlauthor{David S. Greenberg}{cne}
\icmlauthor{Marcel Nonnenmacher}{cne}
\icmlauthor{Jakob H. Macke}{cne}
\end{icmlauthorlist}

\icmlaffiliation{cne}{Computational Neuroengineering, Department of Electrical and Computer Engineering, Technical University of Munich, Munich, Germany}

\icmlcorrespondingauthor{~}{\{david.greenberg, marcel.Nonnenmacher, macke\}@tum.de}

\icmlkeywords{Likelihood-free, Simulation-based inference, ABC, Approximate Bayesian Computation, Amortized Inference, Bayesian Inference, Density Estimation, SNPE, SNL, Masked Flows}

\vskip 0.3in
]



\printAffiliationsAndNotice{} 

\begin{abstract}
How can one perform Bayesian inference on stochastic simulators with intractable likelihoods?
A recent approach is to learn the posterior from adaptively proposed simulations using neural network-based conditional density estimators. However, existing methods are limited to a narrow range of proposal distributions or require importance weighting that can limit performance in practice. Here we present automatic posterior transformation (APT), a new sequential neural posterior estimation method for simulation-based inference. APT can modify the posterior estimate using arbitrary, dynamically updated proposals, and is compatible with powerful flow-based density estimators. It is more flexible, scalable and efficient than previous simulation-based inference techniques. APT can operate directly on high-dimensional time series and image data, opening up new applications for likelihood-free inference.
\end{abstract}

\section{Introduction}
Many applications in science, engineering and economics make extensive use of complex simulations describing the structure and dynamics of the process being investigated.
Such models are derived from knowledge of the mechanisms and principles underlying the data-generating process, and are of critical importance for scientific hypothesis-building and testing.
However, linking complex mechanistic models to empirical measurements can be challenging, since many models are defined implicitly through stochastic simulators (e.g. metabolic models, spiking networks in neuroscience, climate and weather simulations, chemical reaction systems, detector models in high energy physics, graphics engines, population models in genetics, dynamical systems in ecology, complex economic models, \ldots).

For complex data-generating processes such as simulation-based models, it is often impossible to compute the likelihood $p(x|\theta)$ of data $x$ given parameters $\theta$, because it involves intractable integrals, because a simulator's internal states are unavailable or because real-world experiments are involved. Classical approaches to such likelihood-free statistical inference \citep[also known as Approximate Bayesian Computation (ABC), see ][]{sisson2018handbook} do not scale to high-dimensional applications, and typically rely on ad-hoc choices to design summary statistics and distance functions.

Recently, several studies have trained conditional density estimators to perform simulation-based inference, while adaptively tuning the simulations to yield informative data. The techniques fall into two main classes, which seek to directly estimate either the likelihood or the posterior:

\emph{Synthetic likelihood} (SL) methods \cite{wood2010statistical,FanNott_13, TurnerSederberg_14} aim to estimate the likelihood $p(x|\theta)$, and plug the results into an inference procedure (such as MCMC) to compute the posterior. A powerful recent approach, sequential neural likelihood (SNL), trains a neural conditional density estimator \citep[e.g.\ masked flow,][]{papamakarios17maf} to estimate $p(x|\theta)$ for all $x$ and $\theta$ \cite{papamakarios18,lueckmann18}.

\emph{Posterior density estimation} approaches directly target the posterior $p(\theta|x)$ by training a density-estimation neural network from (simulated) data $x$ to $\theta$. This approach does not require additional inference procedures, and thus naturally \emph{amortizes} inference. In addition, it leverages the ability of neural networks to learn informative features from data.

However, posterior estimation faces a key challenge: Drawing simulation parameters from the prior is wasteful, but other, adaptively-chosen proposals require either numerically unstable post-hoc corrections \cite{papamakarios16epsfree} or importance weights \cite{lueckmann17} that increase variance during learning (details in \ref{sec:snde}).

Here we propose Automatic Posterior Transformation (APT), a new sequential neural posterior estimation approach that combines desirable properties of posterior estimation (directly targeting the posterior, amortization, feature learning) and likelihood estimation (flexible proposals, no importance weights or post-hoc corrections).
APT learns a mapping from data to the \emph{true} posterior by maximizing the probability of simulation parameters under the \emph{proposal} posterior.
 By recasting inference as a density ratio estimation problem, it can incorporate powerful flow-based density estimators and use arbitrary, dynamically updated proposals.

We demonstrate the effectiveness and flexibility of APT on a variety of problems. It outperforms previous posterior density estimation methods and scales to high dimensional data, with efficient inference on Lokta-Volterra time series and $10k$-dimensional image data {without} summary statistics.

\begin{figure}[bt!]
\centering
\centerline{\centerline{\includegraphics[width=\linewidth]{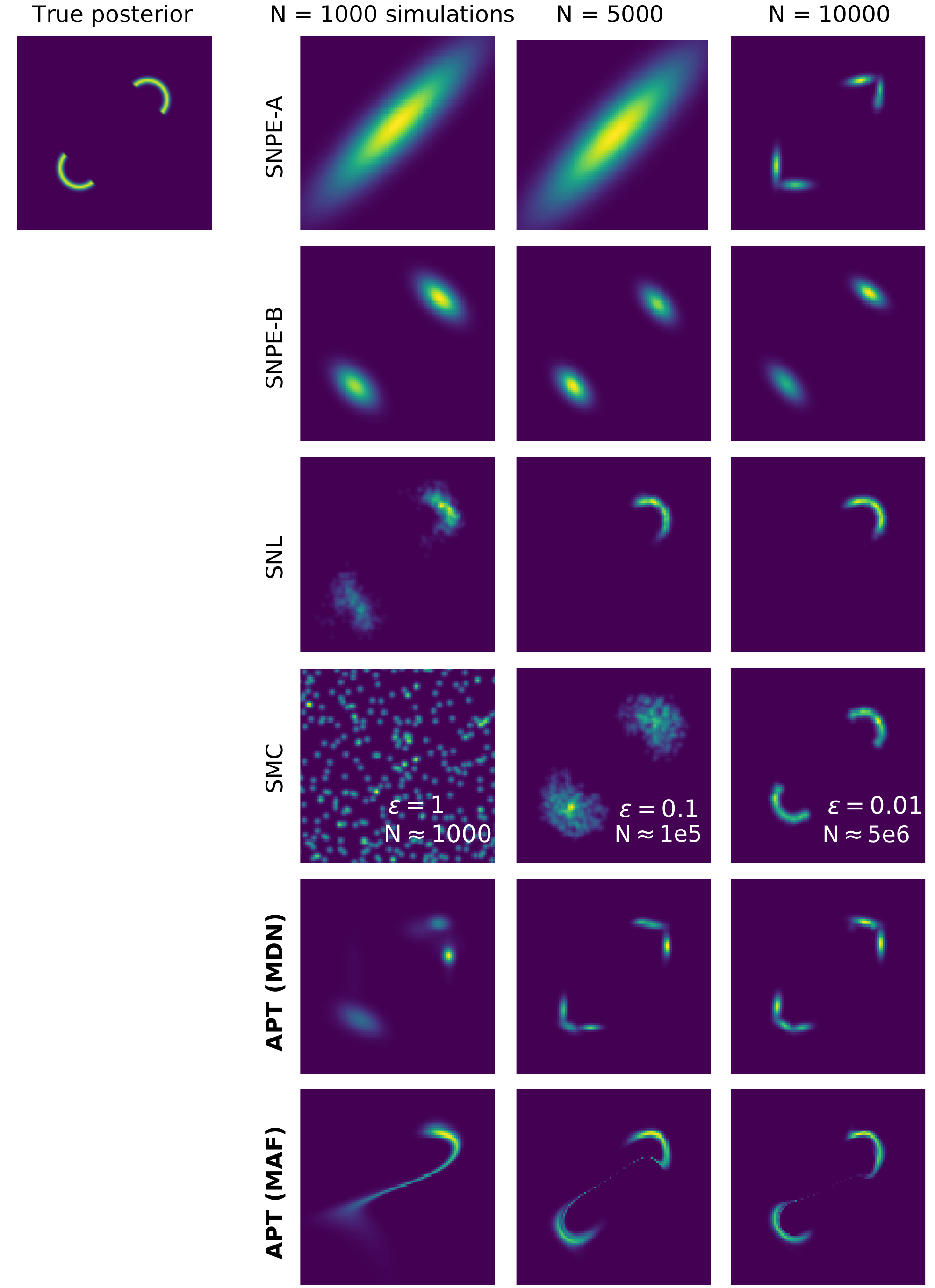}}}
\vskip -0.1in
\caption{\textbf{Comparison of inference algorithms on `two-moons' simulator}. SNPE-A uses Gaussian posteriors/proposals until the last round. SNPE-B's importance weights can lead to slow learning. SNL requires MCMC sampling. SMC-ABC requires a distance function, and far more simulations than other methods. APT with mixture- or flow-based density estimators avoids these limitations.
}
\label{fig:two_moons}
\vskip -0.1in
\end{figure}

\section{Simulation-based inference with conditional density estimators}
\subsection{Problem statement}
Suppose a simulator generates synthetic data $x \in \mathbb R^d$ for any parameter $\theta \in \mathbb R^n$, but the density $p(x | \theta)$ is unknown or intractable. Data $x_o$ are observed from the simulator (or a real-world process it describes) and a prior $p(\theta)$ is known. \textit{Likelihood-free inference} (LFI) seeks to accurately estimate $p(\theta | x_o)$ without access to the likelihood $p(x | \theta)$, by carrying out a limited number of simulations.
\subsection{Conditional density estimation}
Likelihood-free inference can be viewed as conditional density estimation, where the task is to map each simulator result $x$ onto an estimate of the posterior density $p(\theta | x)$. Once the density estimator has been trained on simulated data $x$, it can then be applied to empirical data $x_o$ to compute the posterior. The posterior estimate is selected from a family of densities $q_\psi$, where $\psi$ are distribution parameters. The mapping from $x$ onto $\psi$ is learned by adjusting the weights $\phi$ of a neural network $F$ (or another flexible function approximator) 
so that $q_{F(x, \phi)}(\theta) \approx p(\theta | x)$.

To train the network, we can simulate using prior-drawn parameters to build a dataset $\{(\theta_j, x_j)\}$ and minimize the loss
$\LL(\phi)=-\sum_{j=1}^N \log q_{F(x_j, \phi)}(\theta_j)$ over network weights $\phi$. For sufficiently expressive $F$ and $q_\psi$, the mapping from $x$ to the posterior $p(\theta | x)$ will be learned as $N\rightarrow \infty$. After training, we estimate the target posterior $p(\theta | x_o)$ by $q_{F(x_o, \phi)}(\theta)$.

\subsection{Sequential neural density estimation \label{sec:snde}}
Since we are ultimately interested in the posterior at $x_o$, simulations from parameters with very low posterior density $p(\theta | x_o)$ may not be useful for learning $\phi$. Thus, after initially estimating $\theta|x_o$ using simulations from the prior $p(\theta)$, we want future simulations to use a proposal $\tilde p(\theta)$ which is more informative about $\theta|x_o$ \cite{papamakarios16epsfree,gutmann16bolfi,lueckmann17,sisson2007sequential,BlumFrancois10}. This iterative refinement of the posterior estimate and proposal is known as sequential neural posterior estimation (SNPE).

Unfortunately, minimizing $\LL$ on samples drawn from a proposal $\tilde p(\theta)$ no longer yields the target posterior but rather\footnote{We assume $\tilde p(\theta) = 0$ where $p(\theta) = 0$.}
\begin{align}
\label{proppost}
\tilde p(\theta | x) &= p(\theta | x) \frac{\tilde{p}(\theta) \ p(x)}{p(\theta) \ \tilde p(x)}
\end{align}
where $\tilde p(x) = \int_\theta \tilde p(\theta) p(x | \theta)$. We call $\tilde p(\theta | x)$ the \emph{proposal posterior}. It would be the correct posterior if $\tilde p(\theta)$ were the prior. LFI methods employing neural conditional density estimators differ primarily in how they deal with this problem, with three main approaches developed so far:

SNPE-A \cite{papamakarios16epsfree} trains $F$ to target the proposal posterior $\tilde p(\theta | x)$ and then post-hoc solves (\ref{proppost}) for $p(\theta | x_o)$. To ensure a closed-form solution, $q_\psi$ is restricted to be a mixture of Gaussians (MoG), $\tilde p(\theta)$ to be Gaussian and $p(\theta)$ to be Gaussian or uniform. This approach is simple and can be highly effective, but does not admit multimodal proposals (Fig. \ref{fig:two_moons}, first row, details in \ref{supp:two_moons}). Furthermore, SNPE-A can return non-positive-definite Gaussian covariance matrices when solving (\ref{proppost}).

SNPE-B \cite{lueckmann17} minimizes an importance-weighted loss $-\sum_{j=1}^N \frac{p(\theta_j)}{\tilde p(\theta_j)} \log q_{F(x_j, \phi)}(\theta_j)$. This allows direct recovery of $p(\theta | x)$ from $q_{F(x, \phi)}$ with no correction and no restrictions on $p(\theta)$, $\tilde p(\theta)$ or $q_\psi(\theta)$. However, the importance weights $p(\theta_j) / \tilde p(\theta_j)$ greatly increase the variance of parameter updates during learning, which can lead to slow or inaccurate inference (Fig. \ref{fig:two_moons}, second row).

SNL instead learns a neural conditional density estimate of the likelihood $p(x | \theta)$, allowing simulation parameters to be drawn from any proposal \citep[e.g. the posterior, ][]{papamakarios18} or chosen in an active learning scheme \cite{lueckmann18}. However, instead of directly inferring $p(\theta|x_o)$, it uses additional MCMC sampling that can be costly or inefficient for complex posteriors (Fig. \ref{fig:two_moons}, third row).
Estimating the likelihood (rather than the posterior) can also be more difficult in some cases (see below). Nonetheless, SNL is accurate and efficient on many problems \cite{durkan18}, and generally outperforms classical ABC approaches such as SMC-ABC.

\section{Automatic Posterior Transformation}
APT\footnote{Or SNPE-C in the taxonomy of \cite{papamakarios18}.} is a SNPE technique which combines desirable properties of existing posterior density estimation and likelihood-based approaches. APT learns to infer the \emph{true} posterior by maximizing an estimated \emph{proposal} posterior. It uses (\ref{proppost}) to form a parameterization which makes it possible to automatically transform between estimates of $p(\theta | x)$ and $\tilde p(\theta | x)$, and thus to easily `read-off' the posterior estimate.\footnote{Code available at \url{github.com/mackelab/delfi}.} We will show how this trick avoids the numerical challenges of previous SNPE techniques, and makes it possible to use a wide range of proposals and density estimators.

APT uses $q_{F(x, \phi)}(\theta)$ to represent an estimate of $p(\theta | x)$. To transform this into an estimate of $\tilde p(\theta | x)$, we first observe that by (\ref{proppost}), $\tilde p(\theta|x)\propto p(\theta | x) \tilde p(\theta) / p(\theta)$. We therefore define
\begin{align}
\label{proppostest}
\tilde q_{x,\phi}(\theta) &= q_{F(x, \phi)}(\theta) \frac{\tilde p(\theta) }{p(\theta)}\frac{1}{Z(x, \phi)},
\end{align}
$Z(x, \phi) = \int_\theta q_{F(x, \phi)}(\theta) \frac{\tilde p(\theta) }{p(\theta)}$ is a normalization constant.

We train the network to carry out posterior inference by minimizing $\tilde \LL(\phi)=-\sum_{j=1}^N \log \tilde q_{x,\phi}(\theta_j)$. This procedure recovers both the true and proposal posteriors, as follows:

If the conditional density estimator $q_{F(x, \phi)}$ is expressive enough that some $\phi^*$ exists for which $q_{F(x, \phi^*)}(\theta) = p(\theta | x)$, then by (\ref{proppost}) also $\tilde q_{x,\phi^*}(\theta) = \tilde p(\theta | x)$. Therefore by Prop. 1 of \cite{papamakarios16epsfree}, minimizing $\tilde \LL(\phi)$ yields $q_{F(x,\phi)}(\theta) \rightarrow p(\theta|x)$ and $\tilde q_{x,\phi}(\theta) \rightarrow \tilde p(\theta|x)$ as $N\rightarrow\infty$.

Similar to SNPE-A/B and SNL, APT iteratively refines the network weights $\phi$ and proposal $\tilde p(\theta)$ over multiple simulation rounds. Each round uses a different proposal $\tilde p_r(\theta)$, leading to different transformations in (\ref{proppostest}) when calculating the loss $\tilde \LL$. However, since $q_{F(x,\phi)}(\theta) = p(\theta|x)$ minimizes the expectation $\mathbb E[\tilde \LL]$ for \emph{any} proposal, APT can train on data from multiple rounds simply by adding their loss terms together. In contrast, SNPE-A cannot re-use data across rounds and SNPE-B must apply different importance weights. Algorithm \ref{alg:aptalg} describes APT in the case that each round's final posterior estimate is the next round's proposal.

APT supports a wide range of proposals and density estimators, including mixture-density networks and flows (Fig. \ref{fig:two_moons}, last two rows). Its only requirement is a closed-form solution of (\ref{proppostest}) for $\tilde q_{F(x,\phi)}(\theta)$, to be optimized during learning and sampled from afterwards. We next describe several combinations of $p(\theta)$, $\tilde p(\theta)$ and $q_\psi(\theta)$ for which this is possible.
\begin{algorithm}[th]
   \caption{APT with per-round proposal updates}
   \label{alg:aptalg}
\begin{algorithmic}
   \STATE {\bfseries Input:} simulator with (implicit) density $p(x|\theta)$, data $x_o$, prior $p(\theta)$, density family $q_\psi$, neural network $F(x, \phi)$, simulations per round $N$, number of rounds $R$.
   \STATE
   \STATE $\tilde p_1(\theta) := p(\theta)$
   \FOR{$r=1$ {\bfseries to} $R$}
   \FOR{$j=1$ {\bfseries to} $N$}
   \STATE Sample $\theta_{r, j} \sim \tilde p_r(\theta)$
   \STATE Simulate $x_{r, j} \sim p(x | \theta_{r, j})$
   \ENDFOR
   \STATE $\displaystyle \phi \leftarrow \argmin_\phi \sum_{i=1}^r \sum_{j=1}^N -\log \tilde q_{x_{i, j}, \phi}(\theta_{i, j})$ \COMMENT{using (\ref{proppostest})}
   \STATE $\tilde p_{r+1}(\theta) := q_{F(x_o, \phi)}(\theta)$
   \ENDFOR
   \STATE {\bfseries return} $q_{F(x_o, \phi)}(\theta)$
\end{algorithmic}
\end{algorithm}
\subsection{Gaussian and Mixture-of-Gaussians proposals}
In the simplest case all distributions are Gaussian, with prior $\gdsmall{\theta}{\mu_0}{S_0}$, proposal $\gdsmall{\theta}{\tilde \mu_0}{\tilde S_0}$, posterior $\gdsmall{\theta}{\mu}{S}$ and proposal posterior $ \gdsmall{\theta}{\tilde \mu}{\tilde S}$.
Then (\ref{proppost}-\ref{proppostest}) reduce to
\begin{align}
\label{precisioncorrection}
\tilde S^{-1} &= S^{-1} + \tilde S_0^{-1} - S_0^{-1} \\
\label{meancorrection}
\tilde S^{-1} \tilde \mu &= S^{-1}\mu + \tilde S_0^{-1}\tilde\mu_0 - S_0^{-1}\mu_0
\end{align}
While SNPE-A uses these relations to calculate the true posterior from the proposal posterior \emph{after} learning, APT does the opposite \emph{during} learning. Since the proposal is narrower than the prior in all sensible scenarios, $\tilde S_0^{-1} - S_0^{-1}$ is positive definite and APT cannot produce invalid $\tilde S$.

The updates (\ref{precisioncorrection}-\ref{meancorrection}) can readily be extended to uniform priors by setting $S_0^{-1} = 0$, and to MoG posteriors by summing over components \cite{papamakarios16epsfree}. When $\tilde p(\theta)$ is a MoG with $L$ components and $q_{F(x ,\phi)}(\theta)$ a MoG with $K$ components, solving (\ref{proppostest}) for $\tilde q_{x,\phi}(\theta)$ yields an $LK$-component MoG (\ref{supp:mog_prop}). This allows APT to propose simulation parameters from multimodal distributions.
\begin{table}[ht]
\caption{Properties of posterior inference techniques}
\label{algstable}
\begin{center}
\begin{small}
\begin{sc}
\begin{tabular}{lcccr}
\toprule
Algorithm &  $\tilde p(\theta)$ & $p(\theta)$ &  $q_\psi$ \\
\midrule
SMC-ABC & Any   & Any       & Discrete \\
SNPE-A  & Gauss & Gauss/Uni & MDN\\
SNPE-B  & Any   & Any       & Any \\
SNL     & Any   & Any       & Any (MCMC) \\
APT     & Any   & Any       & Any \\
\bottomrule
\end{tabular}
\end{sc}
\end{small}
\end{center}
\end{table}

\subsection{Atomic proposals}
We would like to extend APT to arbitrary choices of the density estimator, proposal and prior, and especially to powerful flow-based density estimators \cite{rezende2015variational, papamakarios17maf, kingma2018glow}. However, in many cases we cannot solve the integral defining $Z$ in $(\ref{proppostest})$. In this section, we show how APT can be trained using `atomic' proposals that only consider a finite set of parameter vectors (`atoms') for each simulation, replacing integrals by sums. Provided that each simulation's parameters are drawn from a different atomic proposal, and that the overall range of possible atoms covers the posterior support, we can infer the full, continuous posterior.

We set $\tilde p(\theta) = U_\Theta$, where $U_\Theta$ is uniform on $\Theta = \{\theta_1,\ldots,\theta_M\}$. Then $\tilde p(\theta | x)$ and $\tilde q_{x,\phi}(\theta)$ are categorical distributions, $\tilde \LL$ is a cross-entropy loss and (\ref{proppost}-\ref{proppostest}) reduce to
\begin{align}
\label{discreteproppost}
\tilde p(\theta | x) &= \frac{p(\theta | x)/p(\theta)}{\sum_{\theta' \in \Theta} p(\theta'|x)/p(\theta')}
\\
\label{discreteproppostest}
\tilde q_{x,\phi}(\theta) &= \frac{q_{F(x, \phi)}(\theta)/p(\theta)}{\sum_{\theta' \in \Theta} q_{F(x, \phi)}(\theta')/p(\theta')}
\end{align}
For fixed $\Theta$, $\mathbb E_{\theta\sim U_\Theta, x\sim p(x|\theta)}[\tilde \LL]$ is minimized precisely when $\tilde q_{x,\phi}(\theta)=\tilde p(\theta|x)$. In that case $\forall \theta_1, \theta_2\in\Theta$, $\frac{q_{F(x, \phi)}(\theta_1)}{q_{F(x, \phi)}(\theta_2)} = \frac{p(\theta_1 | x)}{p(\theta_2 | x)}$ by (\ref{discreteproppost}-\ref{discreteproppostest})---that is, posterior density \textit{ratios} are correct over all atoms and for every $x$ the atoms can generate.

Now suppose that $\Theta$ is itself sampled from a `hyperproposal' $V(\Theta)$ for each simulation. To minimize $\mathbb E [\tilde \LL]$, the network must then infer the correct posterior density ratios for every pair of samples from each $\Theta$. We formalize this argument as follows (proof in \ref{supp:proof_prop1}):

\begin{proposition}\label{prop1} Let $\Theta\sim V$, and let

$\rho(x, \Theta, \phi) = \begin{cases}
1, & \text{if } \frac{p(\theta_1 | x)}{p(\theta_2 | x)} = \frac{q_{F(x, \phi)}(\theta_1)}{q_{F(x, \phi)}(\theta_2)} \quad\forall \theta_1, \theta_2\in\Theta \\
0, & \text{otherwise}
\end{cases}$

and suppose that for some $\phi^*$, $q_{F(x, \phi^*)} = p(\theta | x)$. Then for any $\phi$ with $\mathbb E_{\Theta\sim V, \theta\sim U_\Theta, x\sim p(x|
\theta)} [\tilde\LL(\phi) - \tilde \LL(\phi^*)] = 0$, %
$\mathbb E_{\Theta\sim V, \theta\sim U_\Theta, x\sim p(x|
\theta)}[\rho(x, \Theta, \phi)] = 1$.
\end{proposition}
By Prop. \ref{prop1}, APT recovers the full posterior shape as $N\rightarrow\infty$, provided that each $\Theta$ is constructed by sampling $\theta$'s independently from a distribution covering the support of $p(\theta | x_o)$. Furthermore, we can get this consistency guarantee using a fixed, nonzero fraction of the atomic proposals to cover the posterior support, while arbitrarily assigning the rest (e.g. for active learning). 

Atomic APT's training data consists of triples $(\theta, x, \Theta)$. $q_{F(x, \phi)}$ is plugged into (\ref{discreteproppostest}) to yield $\tilde q_{x,\phi}$, which is evaluated on $\theta$ to calculate $\tilde \LL$. In practice, instead of sampling $\Theta$, $\theta$ and $x$ during training we simulate once per round as in non-atomic APT, then resample the triples $(\theta, x, \Theta)$ from all rounds' simulations during training. We describe the full algorithm and its computational complexity in \ref{supp:algorithm}.

Learning with such `atomic' proposals has an intuitive interpretation: we are training the network to solve multiple choice test problems, of the format ``which of these $\theta$'s generated this $x$?'' Given a list of $M$ possible alternatives, the network gives a Bayesian answer by assigning mass to each $\theta\in\Theta$. Prop. \ref{prop1} shows we can learn to infer continuous posteriors by training on multiple choice questions.

\begin{figure*}[ht]
\begin{center}
\centerline{\includegraphics[width=\linewidth]{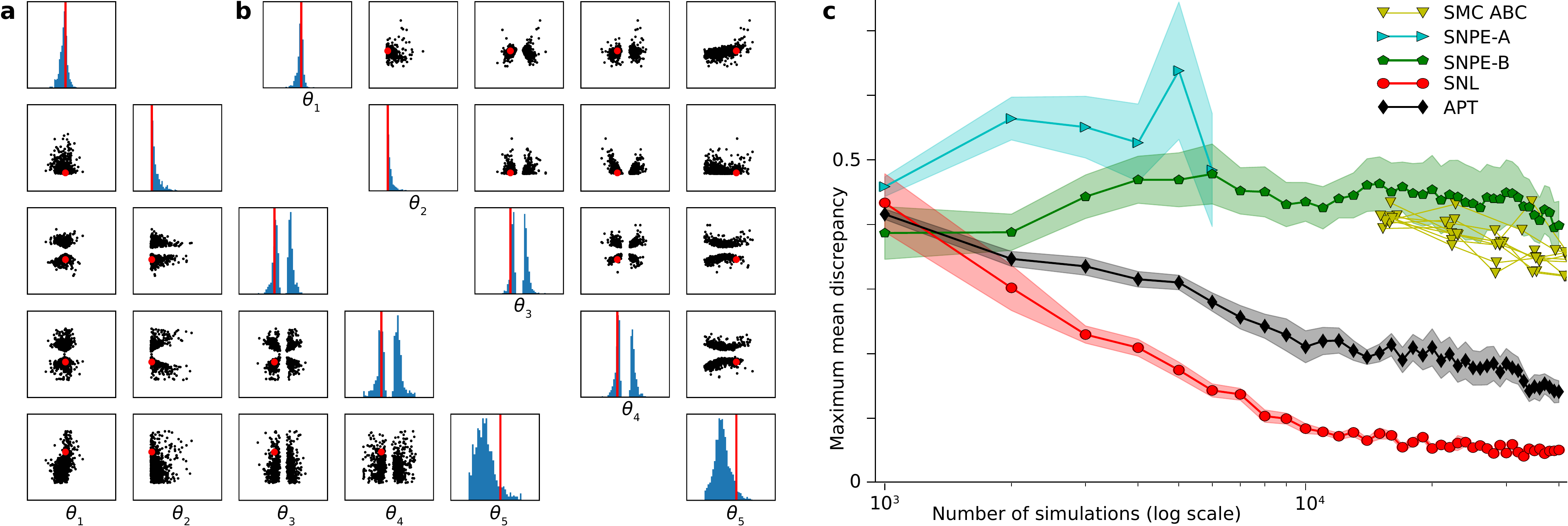}}
\vskip -0.1in
\caption{\textbf{Performance on a model with a simple likelihood and complex posterior (SLCP)}. \textbf{a} Posterior samples from flow-based APT, and \textbf{b} SNL (ground-truth parameters in red).
\textbf{c} Maximum mean discrepancies between estimated and ground-truth posteriors. Mean $\pm$ SEM over $10$ random initializations of the simulator and inference methods with identical $x_o$. SNPE-A terminated after round $6$. APT outperforms previous SNPE methods, closing over half the performance gap to SNL on a problem designed to favor likelihood estimation.
}
\label{fig:gauss}
\end{center}
\vskip -0.3in
\end{figure*}
By replacing integrals with sums, we can use flows or any other distributions for the prior, proposal and posterior estimate, so long as the densities can be easily evaluated, we can sample from the proposal and the posterior is differentiable in $\psi$. Atomic proposals also greatly enhance flexibility when generating new simulations: they can emphasize the posterior's peaks or tails, or use active learning schemes \cite{lueckmann18, jarvenpaa2018efficient} incompatible with previous SNPE approaches.

\subsection{Truncated priors and posteriors}
With atomic proposals and a prior with limited support, Prop. 1 only ensures we can retrieve the posterior $p(\theta | x)$ up to an unknown scale factor: $q_{F(x, \phi)}$ can be trained to match its shape, but not its amplitude. In this case we can easily obtain posterior samples by sampling from $q_{F(x, \phi)}$ and rejecting when $p(\theta)=0$. This is less convenient than directly sampling from the posterior, but still simpler and more efficient than a full MCMC scheme. We discuss the effects of truncated priors on posterior estimation in \ref{supp:prior_truncation}.

\section{Experiments}
We compare APT to SNPE-A, SNPE-B and SNL on several problems (implementation details in \ref{supp:experiments}). \cite{papamakarios18} quantitatively compared these algorithms with classical ABC methods (SMC-ABC and SL).

\subsection{Illustrative toy example with multiple modes\label{sec:two_moons}}
We first illustrate properties and common failure cases of inference methods on a simple `two moons' simulator with 2D parameters $\theta$ and observations $x$ (Fig. \ref{fig:two_moons}, details in \ref{supp:two_moons}, quantitative evaluation in Fig. \ref{fig:two_moons_eval}). Its conditionals $p(\theta|x)$ have both global (bimodality) and local (crescent shape) structure. The posterior was learned over $10$ rounds of $1000$ simulations. We can see clear differences between methods: 

SNPE-A is limited to Gaussian proposals and hence cannot learn fine structure until the final round. SNPE-B has MoG proposals, but its importance weights can lead to slow learning: here it fails to capture the crescent shapes. SNL can flexibly approximate the crescent-shaped likelihood $p(x|\theta)$, but then requires an additional inference step to calculate the posterior from the synthetic likelihood---the coordinate-alternating slice sampling method previously used by \cite{papamakarios18} here fails to mix well between the two non-axis-aligned modes (we emphasize that other MCMC approaches or parameter settings might work better here---however, multimodal and high-dimensional posteriors are generally challenging for MCMC). 
Classical SMC-ABC \cite{sisson2007sequential,BeaumontCornuet_09} requires a large number of simulations (details in \ref{supp:two_moons}; note higher number of simulations used for SMC). 

APT with a Mixture-Density Network (MDN) for density estimation identifies the two-moons structure, and uses it to efficiently guide proposals for subsequent rounds. APT can also be flexibly applied to other density estimators, as illustrated here using a masked autoregressive flow \cite{papamakarios17maf} to represent the posterior.
\subsection{Toy example with simple likelihood but complex posterior}
\cite{papamakarios18} introduce a toy example designed to have a Simple Likelihood and Complex Posterior (SLCP model, details in \ref{supp:gauss}). Its likelihood $p(x|\theta)$ is a Gaussian in $x$ whose mean and covariance depend nonlinearly on $\theta$, while its posteriors $p(\theta|x)$ are non-Gaussian and multi-modal in $\theta$. Thus, estimating the posterior $p(\theta|x)$ is more challenging than estimating the likelihood $p(x|\theta)$. Nevertheless, we demonstrate that with flexible flow-based conditional density estimators, APT's posterior estimates are similar to SNL's \citep[Fig. \ref{fig:gauss}\textbf{a} vs. \textbf{b}, ground-truth posterior in][]{papamakarios18}. A quantitative evaluation \citep[using Maximum Mean Discrepancy,][]{gretton2012kernel} shows that APT outperforms SNPE-A/B, and is closer in performance to SNL, on a problem which is designed to favor SNL.
\begin{figure}[ht]
\begin{center}
\centerline{\includegraphics[width=\linewidth]{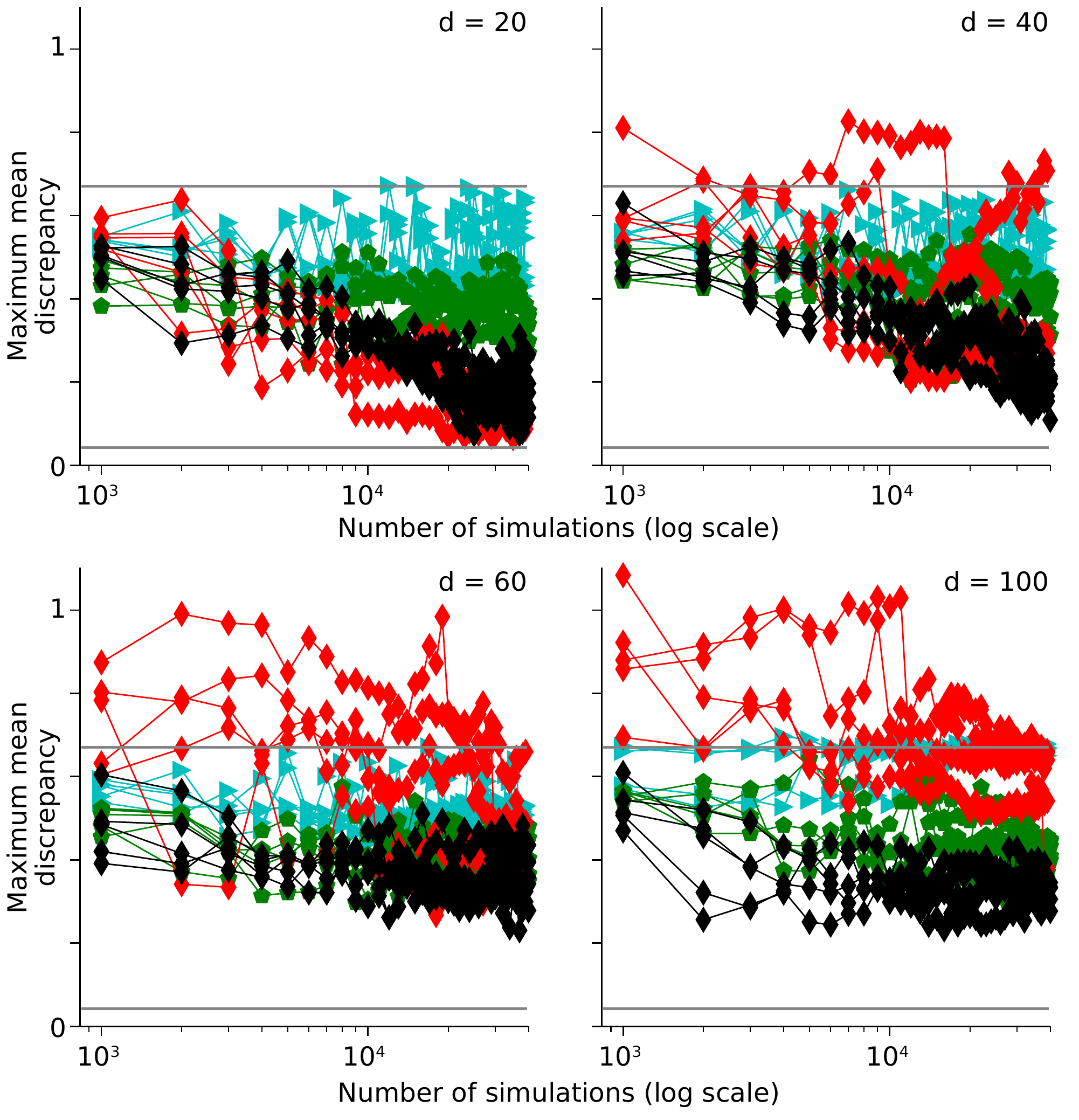}}
\vskip -0.1in
\caption{\textbf{Inference with uninformative data dimensions}.
Appending uninformative dimensions to the SLCP model demonstrates APT can learn relevant features. Maximum mean discrepancies (MMDs) between estimated and ground-truth posteriors ($5$ random initializations) for each value of $d = \dim(x)$. Gray lines show MMDs for prior (upper) and ground-truth posterior samples.
}
\label{fig:noisedims}
\end{center}
\vskip -0.2in
\end{figure}
\subsection{Effect of non-informative observations}
Posterior inference for high-dimensional data is a challenging problem \cite{ong2018likelihood}, especially when multiple data dimensions are uninformative about $\theta$. This scenario is common in neuroscience \cite{paninski2017neural}, physics \cite{brehmer18mining}, systems biology \cite{clarke2008properties} and inverse graphics \cite{romaszko2017vision}.

While classical ABC methods require model-specific low-dimensional summary statistics, APT maps from $x$ to a posterior estimate using deep neural networks, which excel at learning relevant features from data. To test whether this would allow efficient inference in models with many uninformative data dimensions, we appended uninformative outputs drawn from a mixture of Student-t distributions to the 8D SLCP ouputs (details in \ref{supp:gauss}). These noise outputs are generated independently from informative outputs, so the true posterior $p(\theta|x_o)$ is unaffected but inference becomes more difficult as the data dimensionality $d$ increases.

At moderate $d$ both SNL and APT can recover $p(\theta|x_o)$, but at large $d$ SNL's posterior estimates are no better than the prior, while APT degrades only slightly (Fig. \ref{fig:noisedims}). Presumably, SNL `invests' too many resources into estimating densities on irrelevant dimensions. For fixed $\dim(\theta)$, the number of network weights grows linearly with $d$ for posterior density estimators, but quadratically for MDN- or flow-based likelihood estimators.
\subsection{Population ecology model}
The Lotka-Volterra model of coupled predator and prey populations \cite{lotka1920analytical} is a classical likelihood-free inference benchmark.
$\theta\in\mathbb R^4$ governs growth rates and predator-prey interactions, and $x$ consists of population counts (here at $150$ fixed intervals). Simulation results are typically compressed into a set of summary statistics $\bar x$ with $d=9$.

\begin{figure}[h!]
\begin{center}
\centerline{\includegraphics[width=\linewidth]{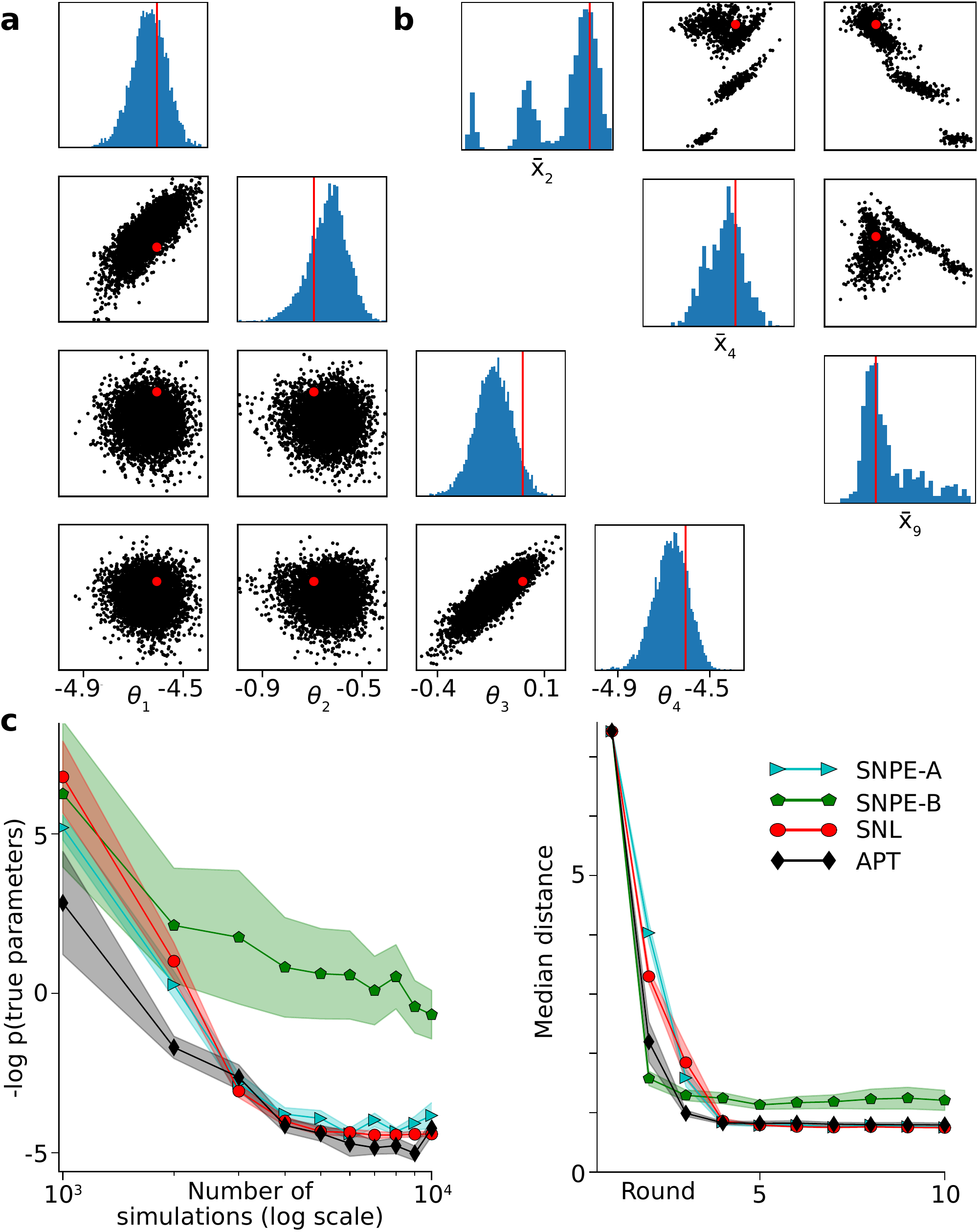}}
\vskip -0.1in
\caption{\textbf{Lotka-Volterra with summary statistics}. The Lotka-Volterra model has simple posteriors and complex conditionals $p(\bar x|\theta)$. \textbf{a} Posterior estimated by APT (ground-truth parameters in red). \textbf{b} Samples from $p(\bar x_2,\bar x_4,\bar x_9|\theta)$ for ground-truth parameters ($\bar x_o$ in red). \textbf{c} Negative log-probabilities of ground-truth parameters and median distances $|\bar x - \bar x_o|$ under the posterior estimates across rounds (means $\pm$ SEM across $10$ different random initializations).
}
\label{fig:lv}
\end{center}
\vskip -0.2in
\end{figure}

This simulator, with complex $\bar x$ distributions (Fig. \ref{fig:lv}\textbf{b}) and simple posteriors (Fig. \ref{fig:lv}\textbf{a}), is well-suited to posterior density estimation. APT produced tight posteriors around the true parameters with fewer simulations than other methods (Fig. \ref{fig:lv}\textbf{c}). While the ground truth posterior is unavailable for this model, for large $N$ APT and SNL compute similar posterior estimates in very different ways, suggesting a close approximation has been obtained (see \ref{supp:lv}).

\subsection{Inference on raw time-series}
\begin{figure}[ht]
\begin{center}
\centerline{\includegraphics[width=\linewidth]{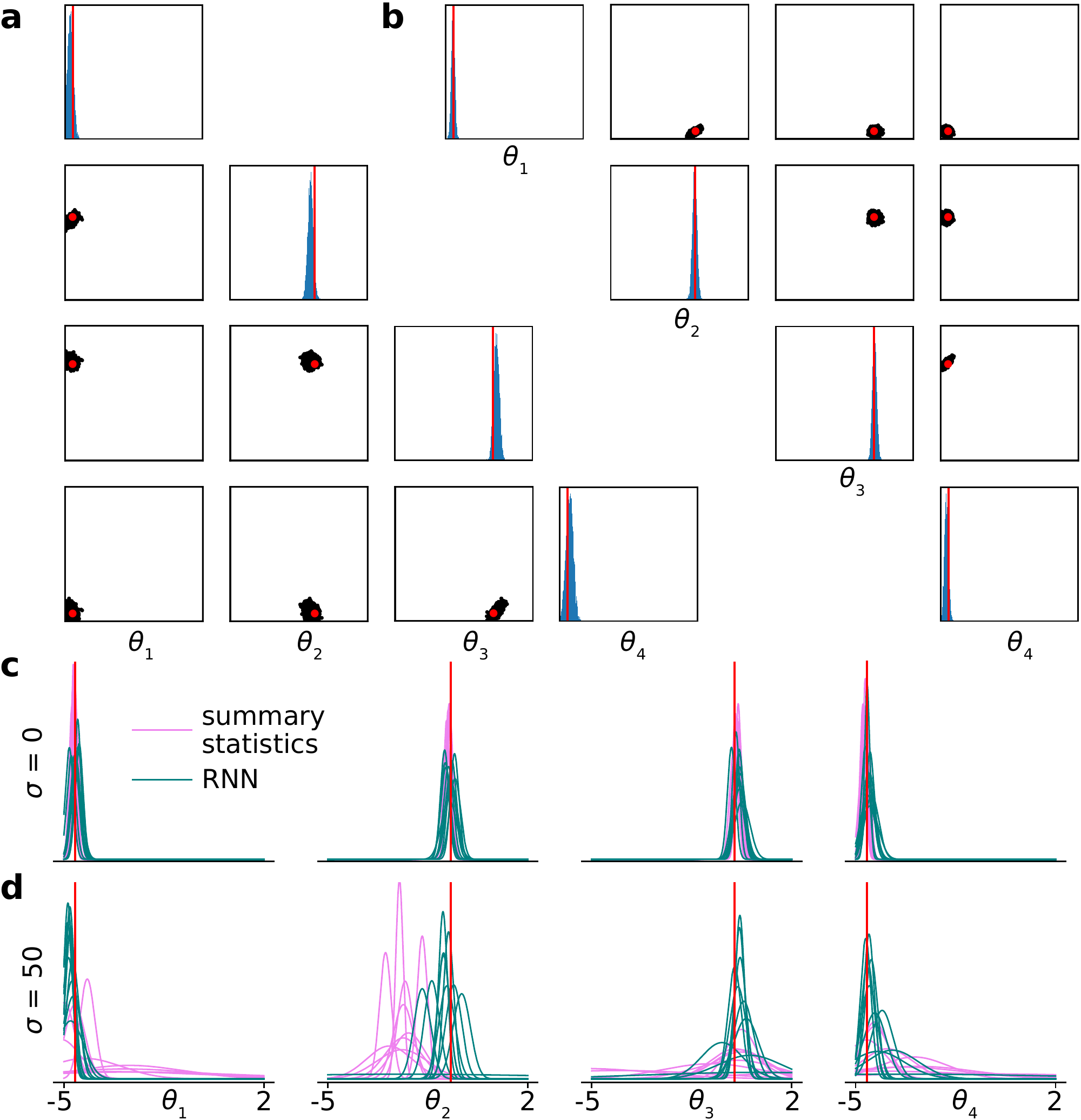}}
\vskip -0.1in
\caption{\textbf{Lotka-Volterra with RNN}.
\textbf{a} Posterior estimate from RNN-based APT (ground-truth parameters in red).
\textbf{b} Posterior estimate from APT with summary statistics (same data).
\textbf{c} Estimated posterior marginals from 10 random initializations, using APT with the RNN (teal) or summary statistics (magenta).
\textbf{d} As in 'c' but with i.i.d. Gaussian observation noise, $\sigma = 50$.
}
\label{fig:rnn}
\end{center}
\vskip -0.3in
\end{figure}

A central problem in ABC is the fact that virtually all algorithms rely on summary statistics in order to reduce the dimensionality of the data \cite{fearnhead2012constructing,blum2013comparative,jiang2017learning}---designing summary statistics can require domain-specific knowledge and/or separate optimization procedures, and might bias algorithms if the statistics do not adequately summarize the informative features of the data. 
To investigate whether sequential neural posterior estimation with APT can be applied to raw simulator outputs, we also applied APT using a recurrent neural network (RNN) to directly map $x$ onto a posterior density estimate (details in \ref{supp:lv}). RNN-APT's estimate of $p(\theta | x)$ (Fig. \ref{fig:rnn}\textbf{a}) was nearly identical to the previous $p(\theta | \bar x)$. Over repeated runs of the algorithm the posterior marginals remained tightly clustered around the true parameters in both cases (Fig. \ref{fig:rnn}\textbf{c}). This shows that APT can operate without summary statistics on high-dimensional data, and suggests that $p(\theta | \bar x)$ may be close to $p(\theta | x)$.

We also examined a more difficult version of the same problem that has been used as an LFI benchmark \cite{owen2015likelihood}, where the observed counts are corrupted by observation noise (Fig. \ref{fig:rnn}\textbf{d}). In this case posterior distributions inferred by the RNN were broader, but still clustered around the true parameters. Posteriors inferred from summary statistics were significantly broader and farther from the true parameters than those of the RNN, suggesting that the summary statistics may discard information about the parameters in this case. By training APT end-to-end we were able to automatically extract relevant information from the data, without using bespoke summary statistics.
\subsection{Rock-paper-scissors reaction-diffusion model}
Simulators that produce images pose a particular challenge for inference, as they produce high-dimensional data without well-tested summary statistics. In evolutionary game theory, rock-paper-scissors (RPS) models \cite{may1975nonlinear} describe nontransitive predator-prey interactions in which species A preys on B, B preys on C and C preys on A, often exhibiting stable biodiversity over long timescales. When each species' population density varies over a 2-dimensional space, the resulting stochastic partial differential equation (SPDE) exhibits complex, dynamically shifting spatial structure \cite{reichenbach2007mobility}. The system's stability and structure depend on its growth, predation and diffusion rates.

\begin{figure}[h]
\begin{center}
\centerline{\includegraphics[width=\linewidth]{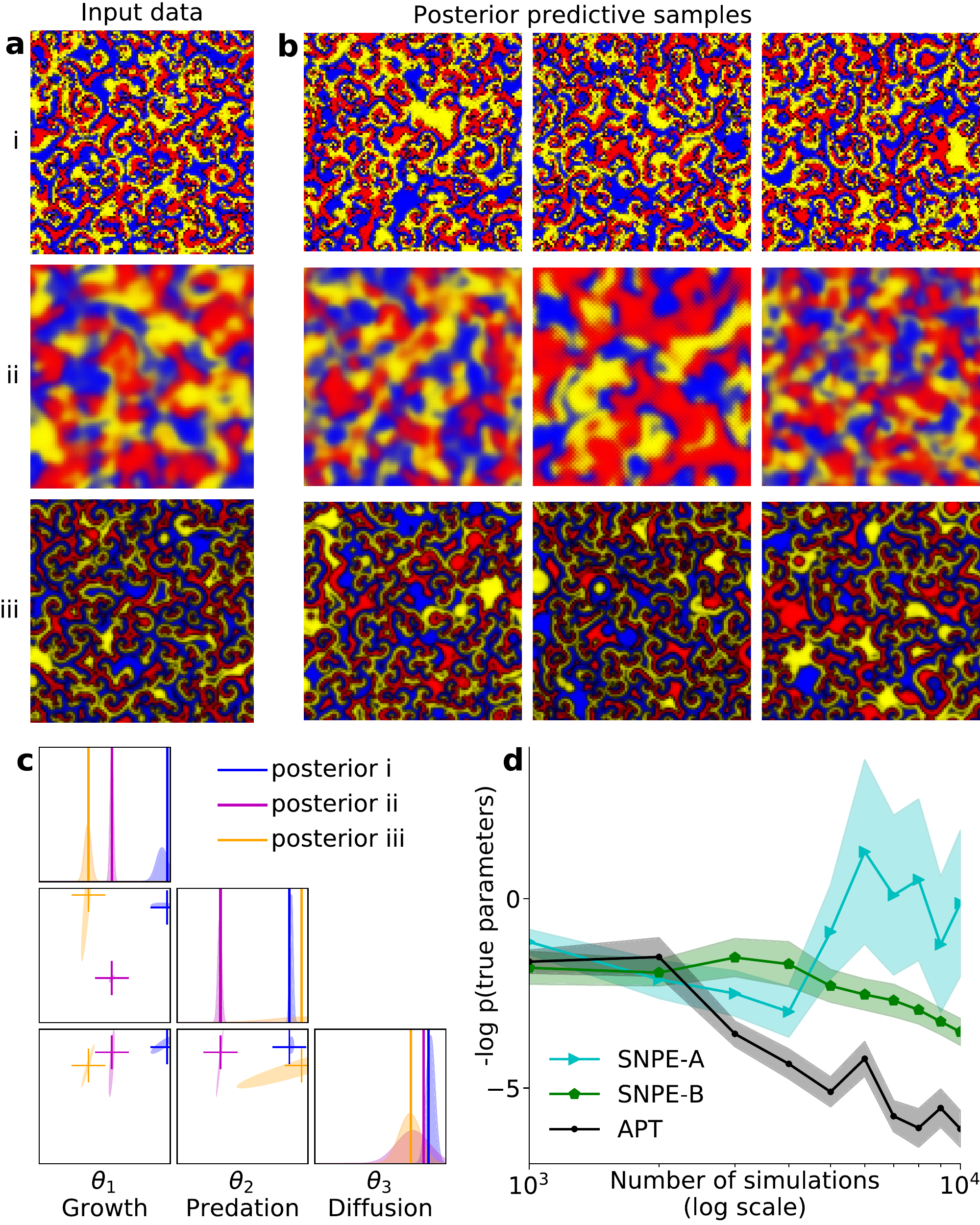}}
\vskip -0.1in
\caption{\textbf{Rock-paper-scissors model with CNN}.	
\textbf{a} Observations $x_o$ for $3$ RPS parameter sets.
\textbf{b} $3$ APT posterior predictive samples $x$ for each $x_o$ in `a': $\theta \sim q_{F(x_o,\phi)}(\theta), x \sim p(x|\theta)$.
\textbf{c} APT posteriors for each $x_o$ in `a.' On-diagonal graphs show marginals; off-diagonals show 2 s.d.s around the mean. Solid lines and crosses show ground truth parameters.
\textbf{d} Negative log-probabilities of ground-truth parameters; mean $\pm$ SEM over $30$ inference problems with ground truth parameters drawn from the prior.
}
\label{fig:rps}
\end{center}
\vskip -0.3in
\end{figure}

We simulated from the SPDE-RPS model, spatially discretized on a 100 x 100 lattice and initialized at the unstable uniform steady state. The observation $x_o \in \mathbb R^{10000}$ was the final system state, which can be displayed as a 3-channel image (Fig. \ref{fig:rps}\textbf{a}). To test whether APT could identify relevant features and infer parameter posteriors from high-dimensional image data, we used a convolutional neural network (CNN) as $F(x, \phi)$. APT inferred posteriors that closely encompassed the ground truth parameters used to generate each $x_o$ (Fig. \ref{fig:rps}\textbf{c}), and posterior-drawn parameters produced simulations that visually resembled $x_o$ (Fig. \ref{fig:rps}\textbf{b}). Running SNPE-A/B with the same CNN yielded lower posterior probability for the ground truth parameters (details in \ref{supp:rps}).
\subsection{Additional experiments}
Finally, we also evaluated two additional benchmark models. On the M/G/1 model \cite{papamakarios16epsfree}, APT outperformed SNPE-A/B and approached SNL in performance (Fig. \ref{fig:mg1_supp}). On the GLM model \cite{lueckmann17} APT was slightly more accurate than SNPE-A, while SNL and SNPE-B were considerably less accurate (Fig. \ref{fig:glm_supp}).
\section{Related work}
Classical ABC approaches \cite{sisson2018handbook}, are based on simulating data from proposal distributions \cite{PritchardSeielstad_99,beaumont2002approximate,MarjoramMolitor_03,BeaumontCornuet_09,gutmann16bolfi} and accepting samples that are sufficiently similar to the observed data. However, this approach generally requires choosing low-dimensional summary statistics, distance functions and rejection thresholds, and is exact only in the limit of high rejection rates. Several papers have investigated procedures for (semi) automatically designing summary statistics \cite{fearnhead2012constructing,blum2013comparative,jiang2017learning, dinev2018dynamic}.

Approaches based on estimating the likelihood \cite{wood2010statistical,FanNott_13, TurnerSederberg_14,papamakarios18,lueckmann18} or posterior \cite{le2017modelbased,papamakarios16epsfree,lueckmann17,chan18} \citep[which can be traced back to regression-adjustment,][]{beaumont2002approximate, BlumFrancois10} are reviewed above (\ref{sec:snde}). Posterior inference with end-to-end feature learning has previously been demonstrated with RNNs \cite{lueckmann17}
and exchangeable networks \cite{chan18}. Some LFI methods instead aim to infer a point or local estimate of the parameters \cite{pesah18, louppe2017adversarial, mccarthy2017variational}.

Atomic APT trains the posterior model through a series of multiple-choice problems, by minimizing the cross-entropy loss from supervised classification.
Several recent studies investigated the use of (neural) classifiers for hypothesis testing or estimation of likelihood ratios: \cite{dutta2016likelihood,gutmann2018likelihood} train binary classifiers to discriminate between the conditional and the marginal distribution, \cite{cranmer2015approximating} trains classifiers to approximate likelihood ratios for frequentist parameter estimation and hypothesis testing and \cite{brehmer18mining} presents approaches for learning likelihoods (and scores) from simulator outputs. \cite{tran17implicit} trains a log ratio estimator to minimize a variational objective, extending neural posterior estimation to implicit density families with intractable $q_\psi(\theta)$, but uses prior-sampled parameters and requires training two networks simultaneously. Atomic APT differs from these in using a single parameterized posterior estimate to derive a different classifier for each multiple choice problem.

If the simulator also provides access to additional information (e.g. internal variables and their likelihoods or an unbiased estimate of $p(x|\theta)$), this information can be exploited to improve efficiency \cite{tran17implicit, brehmer18mining,andrieu2010particle,tran2017variational}. Approaches for so-called \emph{implicit} generative models generally require the simulator to be differentiable with respect to model parameters \citep[][and references therein]{Mohamed_16,Huszar_17}. While generative adversarial networks (GANs) also train classifiers on simulated data, our learning objective is different \cite{Huszar_17}, as in particular as we have no `adversary.' Nevertheless, adversarial training can also be extended to inference problems, and its application to likelihood-free inference has been proposed \cite{louppe2017adversarial}. It might be possible to combine adversarial learning with APT, especially when using atomic proposals.
\section{Discussion}
Density estimation is an attractive approach to simulation-based Bayesian inference. We presented APT, which can be applied to arbitrary proposals and a wide range of conditional density estimators. Flexibility in choosing proposals opens up multiple new applications and opportunities for extensions, e.g.\ combining posterior estimation with active-learning rules, or loss-calibrated likelihood-free inference. APT is more simulation-efficient than previous posterior density estimation approaches on a variety of problems, and can scale to higher dimensional observations.

Depending on the model and analysis problem at hand, the likelihood or the posterior might be easier to approximate. We do not claim that directly targeting the posterior will always outperform synthetic-likelihood approaches. However, it has the advantage of directly yielding a mapping from data to posterior without an additional inference step, and therefore amortizes inference. An advantage of targeting the posterior directly is that specialized neural network architectures can be used to exploit known structure in the data, as we have shown using RNNs and CNNs for time series and image data. Exchangeable neural networks \cite{zaheer2017deep, chan18, bloem2019probabilistic} could also be applied when $x_o$ consists of multiple i.i.d. observations from the same parameters. 

We hope that the combination of flexible conditional density estimators, effective schemes for adaptively choosing simulations, and stable learning frameworks will eventually make statistical inference efficient, easy and (ideally) automated, for a wide range of simulation-based models.
\clearpage
\section*{Acknowledgements} We thank Jan-Matthis Lueckmann, Álvaro Tejero-Cantero, Poornima Ramesh, Pedro J. Gonçalves, Jan Bölts, Michael Deistler and Artur Speiser for comments and discussions, Poornima Ramesh for pointing us to the lattice RPS model, Tobias Reichenbach for RPS code and the anonymous reviewers for detailed and constructive feedback. Funded by the German Research Foundation (DFG) through SFB 1233 (276693517), SFB 1089 and SPP 2041 and the German Federal Ministry of Education and Research (BMBF, project `ADMIMEM', FKZ 01IS18052 A-D).
\bibliography{apt}
\bibliographystyle{icml2019}
\clearpage
\appendix
\onecolumn
\section{Supplementary material}
\subsection{Derivation of MoG proposals for APT with MDNs}
\label{supp:mog_prop}
When $q_{F(x,\phi)}$ is an $M$-component mixture of Gaussians, $\tilde p(\theta)$ is an $L$-component mixture of Gaussians and $p(\theta)$ is Gaussian, we have
\begin{align}
q_{F(x,\phi)}(\theta) &= \sum_{i=1}^M \alpha_i
\gd{\theta}{\mu_i}{\Sigma_i} \\
\tilde p(\theta) &= \sum_{k=1}^L \beta_k \gd{\theta}{\tilde \mu_k}{\tilde \Sigma_k} \\
p(\theta | x) &= \gd{\theta}{\mu_0}{\Sigma_0} \\
\tilde q_{x,\phi}(\theta) &= \frac{1}{Z_{x, \phi}}\sum_{i, k}\alpha_i\beta_k \frac{\gd{\theta}{\mu_i}{\Sigma_i} \gd{\theta}{\tilde \mu_k}{\tilde \Sigma_k}}{\gd{\theta}{\mu_0}{\Sigma_0}} \\
&= \sum_{i, k} \zeta_{ik} \gd{\theta}{\mu^*_{ik}}{\Sigma^*_{ik}}
\end{align}
where
\begin{align}
\Sigma^*_{ik} &= \left(\Sigma_i^{-1} + \tilde \Sigma_k^{-1} - \Sigma_0^{-1}\right)^{-1} \\
\mu^*_{ik} &= \Sigma^*_{ik} \left(\Sigma_i^{-1}\mu_i + \tilde \Sigma_k^{-1}\tilde\mu_k - \Sigma_0^{-1}\mu_0\right) \\
\zeta_{ik} &\propto 
\alpha_i \beta_k \sqrt{\frac{\det(\Sigma^*_{ik})}{\det(\Sigma_i)\det(\tilde{\Sigma}_k)}}
e^{-\frac{1}{2}\left(\mu_i^\top \Sigma_i^{-1} \mu_i + \tilde{\mu}_k^\top \tilde{\Sigma}_k^{-1} \tilde{\mu}_k - {\mu_{ik}^*}^\top {\Sigma_{ik}^*}^{-1} \mu_{ik}^*\right)}
\end{align}
and the proportionality symbol indicates that the weights $\zeta_{ik}$ should be normalized so that $\sum_{ik} \zeta_{ik} = 1$.
\clearpage%
\subsection{Algorithm and computational complexity for atomic APT}
\label{supp:algorithm}
\begin{algorithm}[hb]
   \caption{APT with atomic proposals}
   \label{alg:atomicaptalg}
\begin{algorithmic}
   \STATE {\bfseries Input:} simulator with (implicit) density $p(x|\theta)$, data $x_o$, prior $p(\theta)$, density family $q_\psi$, neural network $F(x, \phi)$, simulations per round $N$, number of rounds $R$, number of atoms $M$.
   \STATE
   \STATE $\tilde p_1(\theta) := p(\theta)$
   \STATE $c \leftarrow 0$ \COMMENT{total simulation count}
   \FOR{$r=1$ {\bfseries to} $R$}
   \FOR{$j=1$ {\bfseries to} $N$}
   \STATE $c \leftarrow c + 1$
   \STATE Sample $\theta_{c} \sim \tilde p_r(\theta)$
   \STATE Simulate $x_{c} \sim p(x | \theta_{c})$
   \ENDFOR
   \STATE $V_r(\Theta) := \begin{cases}
{c\choose M}^{-1}, & \text{if } \Theta=\{\theta_{b_1},\theta_{b_2},\ldots,\theta_{b_M}\}\text{ and } 1 \leq b_1 < b_2 < \ldots < b_M \leq c \\
0,   & \text{otherwise}
\end{cases}$ \COMMENT{sampling without replacement}
   \STATE $\phi \leftarrow \argmin_\phi \mathbb E_{\Theta\sim V_r(\Theta)} \left[\sum\limits_{\theta_j\in\Theta} -\log \tilde q_{x_j, \phi}(\theta_j)\right]$
   \STATE $\tilde p_{r+1}(\theta) := q_{F(x_o, \phi)}(\theta)$
   \ENDFOR
   \STATE {\bfseries return} $q_{F(x_o, \phi)}(\theta)$
\end{algorithmic}
\end{algorithm}
\vspace{-0.1in}
To minimize $\mathbb E_{\Theta\sim V_i(\Theta)} \left[\sum\limits_{\theta_j\in\Theta} -\log \tilde q_{x_j, \phi}(\theta_j)\right]$, we must calculate its gradient with respect to $\phi$, for which we use minibatches of size $M$. Specifically, for each minibatch we first sample $B = \{b_1,\ldots,b_M\} \subset \{1, \ldots, c\}$ without replacement. We then calculate the gradient $\frac{d}{d\phi} \sum\limits_{b \in B} -\log \tilde q_{x_b, \phi}(\theta_b)$ using (\ref{discreteproppostest}). Note that each minibatch involves $M$ loss evaluations---that is, $M$ multiple-choice questions with $M$ possible answers each.

To calculate $\tilde q_{x_j, \phi}$ and its gradients for a single minibatch using (\ref{discreteproppostest}), we must evaluate $q_{F(x,\phi)}(\theta)$ on every possible pair $(\theta_b, x_{b'})$ for $b, b' \in B$. Therefore atomic APT's computational complexity is quadratic in the minibatch size $M$. However, we observed no difference in wallclock time per minibatch for $M=10$ vs. $M=100$ on an nVidia GeForce RTX 2080. Furthermore, for the Lokta-Volterra and RPS models, simulations took longer than all other calculations, for all methods.

When using an MDN as the density estimator, the MoG estimate of the posterior can be calculated once for each $x_{b'}$, and then each MoG evaluated on each $\theta_b$. For a network with $n_\text{layers}$ fully connected hidden layers of $n_\text{hidden}$ units each, $d = \dim(x)$ and an $n_\text{MoG}$-component Gaussian mixture, the computational complexity of an atomic APT minibatch is
\begin{align}
C_\text{atomic MDN-APT} =
\mathcal O\left(M \left [d n_\text{hidden} + n_\text{layers} n_\text{hidden}^2 + n_\text{hidden} n_\text{MoG}\dim(\theta)^2\right]
+ M^2 n_\text{MoG}\dim(\theta)^2 \right)
\end{align}
Note that only the final term is quadratic in $M$, and this term does not involve the network structure or input dimensionality. For comparison, SNPE-A/B or non-atomic MDN-based APT has the same complexity, except for being linear in $M$:
\begin{align}
C_\text{SNPE-A/B/non-atomic MDN-APT} =
\mathcal O\left(M\left[d n_\text{hidden} + n_\text{layers} n_\text{hidden}^2 + n_\text{hidden} n_\text{MoG} \dim(\theta)^2\right] \right)
\end{align}
In a MAF data and parameters are coupled, so for atomic APT every $(\theta_b, x_{b'})$ pair requires a separate feedforward pass. For $n_\text{MADEs}$ conditional MADEs consisting of $n_\text{layers}$ fully connected hidden layers of $n_\text{hidden}$ units, the minibatch complexity is
\begin{align}
C_\text{atomic MAF-APT} =
\mathcal O\left(n_\text{MADEs}\left [
M d n_\text{hidden} + M ^ 2 \dim(\theta) n_\text{layers} n_\text{hidden}^2 + M^2 \dim(\theta)^2 n_\text{hidden}
\right]\right)
\end{align}
The first term is exempted from quadratic dependence on $M$ since the inputs $x$ can be multiplied by the appropriate weight matrices independently of $\theta$. Thus, any computational tasks that scale with the input dimension $d$ scale only linearly with $M$.

For MAF-based SNL the roles of the data and parameters are reversed, so the complexity (not including MCMC sampling) is
\begin{align}
C_\text{MAF-SNL} =
\mathcal O\left(M n_\text{MADEs}\left [
\dim(\theta) n_\text{hidden} + d n_\text{layers} n_\text{hidden}^2 + d^2 n_\text{hidden}
\right]\right)
\end{align}
For very large $d$, we hypothesize that atomic MAF-APT could benefit from specialized network architectures that pass $x$ through a learned, feed-forward dimensionality reduction before supplying the result as input to all MADEs.
\clearpage%
\subsection{Conditional flow normalization and truncated priors}
\label{supp:prior_truncation}

Conditional density estimators trained to target the posterior density should respect constraints imposed on them by the prior.
For uniform priors, the posterior density outside the prior support should be zero. 
The proposal correction for SNPE-A \cite{papamakarios16epsfree} does not hold for tight priors that clearly truncate the posterior estimate.
Also for SNPE-B, posterior leakage leads to hard-to-interpret results unless the MDNs are re-normalized. 

The normalization of the conditional density model $q_{F(x,\phi)}(\theta)$ cancels out in the computation of the probabilities $\tilde q_{x,\phi}(\theta)$ (see eq. \ref{discreteproppostest}). 
Hence conditional density models optimized via APT with atomic proposals are automatically normalized during training.
After training, the Gaussian (mixture) $q_{F(x_o,\phi)}(\theta)$ returned from MDNs can be truncated to return a valid truncated Normal posterior estimate.
For MAFs we can effectively obtain such post-hoc truncation through rejection sampling, and estimate the normalization factor from the rejection rates.

We noticed that across many rounds, conditional MAFs trained with APT can leak increasingly large amounts of mass outside the prior support. We did not find this leakage to negatively influence the quality of the estimated posterior shape (evaluated over the prior support).
If needed, there would be several options to reduce leakage across rounds and keep rejection rates low: We can periodically reinitialize the conditional density estimator across rounds. 
Alternatively, one could also train a new flow which is optimized to have the same 
shape on the prior's support, but minimal mass elsewhere. This normalized flow could then be used to directly evaluate posterior densities.
For simple box-shaped prior supports, one can also apply a pointwise (scaled) logistic transformation to the MAF outputs to enforce prior bounds, i.e. train $q_{F(x,\phi)}(\sigma^{-1}(\theta))$. 

\subsection{Proof of proposition 1}
\label{supp:proof_prop1}
Here we prove Prop. \ref{prop1}. Note that whenever we refer to $\tilde p(\theta)$, $\tilde p(\theta | x)$ or $\tilde p(x)$ these distributions are always defined based on some specific choice of $\Theta$.

\textbf{Proof of proposition \ref{prop1}}
\begin{align}
\mathbb E_{\Theta\sim V, \theta\sim U_\Theta, x\sim p(x|
\theta)}[\tilde\LL(\phi) - \tilde\LL(\phi^*)] &=
\int_\Theta V(\Theta) \sum_{\theta \in \Theta} \tilde p(\theta) \int_x p(x | \theta) [\log \tilde p(\theta | x) - \log \tilde q_{x,\phi}(\theta)]
\end{align}
By Bayes' rule and (\ref{proppost}), for $\theta \in \Theta$ we have $p(x | \theta) = \frac{\tilde p(\theta | x) \tilde p(x)}{\tilde p(\theta)}$ so
\begin{align}
\mathbb E_{\Theta\sim V, \theta\sim U_\Theta, x\sim p(x|
\theta)}[\tilde\LL(\phi) - \tilde\LL(\phi^*)]&=
\int_\Theta V(\Theta) \int_x \tilde p(x) 
\sum_{\theta \in \Theta} \tilde p(\theta | x) \log \frac{\tilde p(\theta | x)}{\tilde q_{x,\phi}(\theta)}
\\&= \int_\Theta V(\Theta) \int_x \tilde p(x) D_\text{KL}(\tilde p(\theta | x) || \tilde q_{x,\phi}(\theta))
\\&= \mathbb E_{\Theta\sim V, \theta\sim U_\Theta, x\sim p(x|
\theta)} [D_\text{KL}(\tilde p(\theta | x) || \tilde q_{x,\phi}(\theta))]
\end{align}
By Gibbs' inequality, the KL divergence is zero only when $\tilde q_{x,\phi}(\theta) = \tilde p(\theta | x)$ for all $\theta\in\Theta$, in which case by (\ref{discreteproppost}-\ref{discreteproppostest}) $q_{F(x, \phi)} \propto p(\theta | x)$ for $\theta\in\Theta$ as well. Thus $D_\text{KL}(\tilde p(\theta | x) || \tilde q_{x,\phi}(\theta)) > 0$ whenever $\rho(x, \Theta, \phi) \neq 1$, so if $\mathbb E_{\Theta\sim V, \theta\sim U_\Theta, x\sim p(x|
\theta)}[ \rho(x, \Theta, \phi)]$ were less than one, $\mathbb E_{\Theta\sim V, \theta\sim U_\Theta, x\sim p(x|
\theta)}[\tilde\LL(\phi) - \tilde\LL(\phi^*)]$ would be greater than zero.

\subsection{Experimental details}
\label{supp:experiments}

We use the same basic network architectures for all of our experiments. 
For mixture-density networks (SNPE-A, SNPE-B, APT), we use two fully-connected tanh layers with with $50$ units each. Unless otherwise stated, we use MDNs with $8$ Gaussian mixture components.
In our experiments with MDNs MoG proposals were used for APT.
For conditional masked autoregressive flows (SNL, APT), we use stacks of $5$ MADEs each constructed using two fully-connected tanh layers with $50$ units each. We train the APT MAFs with atomic proposals using $M=100$ atoms.

For the SLCP, Lotka-Volterra and M/G/1 model, we follow the experimental setup of \cite{papamakarios18}, including uniform priors, summary statistics, ground-truth parameters $\theta^*$ and observed data $x_o$.

\subsubsection{Two moons model}
\label{supp:two_moons}

For given parameter $\theta\in\mathbb{R}^2$, the `two moons' simulator generates $x\in\mathbb{R}^2$ according to
\begin{align}
a &\sim U(-\frac{\pi}{2}, \frac{\pi}{2}) \\
r &\sim \mathcal{N}(0.1, 0.01^2) \\
p &= (r \cos(a) + 0.25, \ r \sin(a)) \\
 x^\top &= p + \left(-\frac{|\theta_1+\theta_2|}{\sqrt{2}}, \ \frac{-\theta_1+\theta_2}{\sqrt{2}} \right) 
\end{align}
The intermediate variables $p$ follow a single crescent-shaped distribution, which is subsequently shifted and rotated around the origin depending on $\theta$. Consequently, $p(x|\theta)$ shows a single shifted crescent for fixed $\theta$.
The absolute value $|\theta_1+\theta_2|$ gives rise to the second crescent in the posterior. We choose a uniform prior over $[-1,1]^2$ to illustrate inference on this model. As observed data we use $x_o = (0,0)^\top$.

On this example we fit mixture-density networks with $20$ mixture components to allow expressive conditional densities.

\begin{SCfigure}[][h]
    \centering
    \includegraphics[width=3.25in]{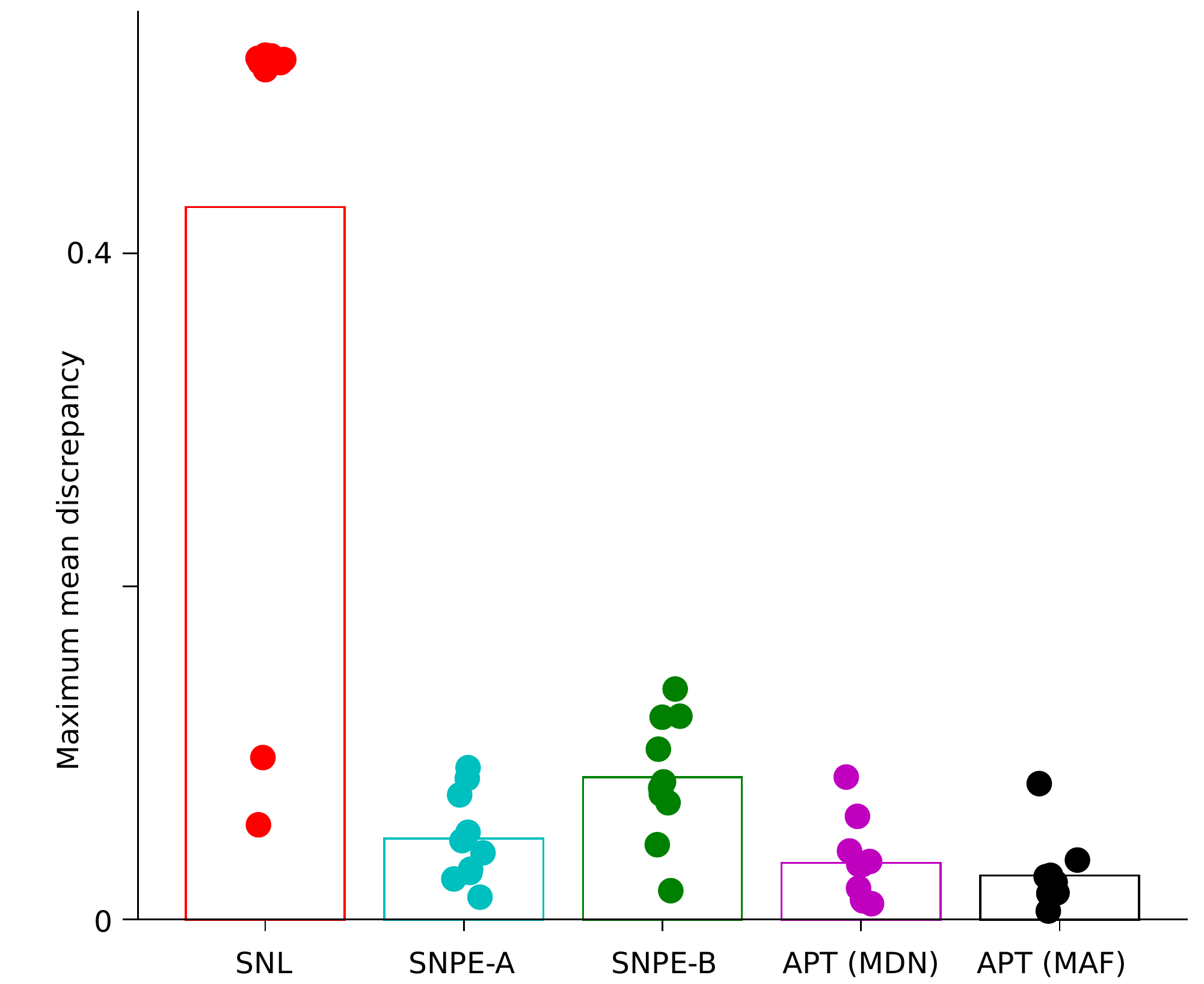}
    \label{fig:two_moons_eval}
    \caption{Two moons model. Comparison of average maximum mean discrepancies between final posterior estimate and ground-truth posteriors for different algorithms across $10$ different random initializations and $x = (0, 0)$. Dots show individual runs. The MCMC chains used by SNL to obtain posterior estimates failed to sample both posterior modes in the majority of cases (cf. Fig \ref{fig:two_moons} for such an example case), leading to high discrepancies. }
\end{SCfigure}

\subsubsection{SLCP model}
\label{supp:gauss}

The SLCP model has a simple simulator density $p(x|\theta) = \prod_{i=1}^4 \mathcal{N}(x_{(2i-1,2i)} | \mu(\theta), \Sigma(\theta))$,
i.e. $x \in \mathbb{R}^8$ consists of four independent samples from a bivariate Gaussian parameterized by $\theta \in \mathbb{R}^5$. 
The conditional mean is given by $\mu(\theta) = (\theta_1, \theta_2)^\top$.
The parameterization of the covariance
\begin{align}
\Sigma(\theta) =  \left[ \begin{array}{c c}
\theta_3^2 & \theta_3 \theta_4 \tanh    (\theta_5) \\
\theta_3 \theta_4 \tanh(\theta_5) & \theta_4^2
\end{array}\right]
\end{align}
leads to in total four modes in $p(\theta|x)$ visible in the pairwise marginal over $(\theta_3, \theta_4)$ (Fig. \ref{fig:gauss}\textbf{a}). To calculate MMD's, We sampled from the ground-truth posterior using MCMC.
For the experiment with added uninformative simulator outputs, we generate noise outputs $x_{i>8} \in \mathbb{R}^{m}, m \in \{12, 32, 52, 92\}$ from $m$-dimensional mixtures of t-distributions and append them to the $8$-dimensional simulator output.
We use mixtures of $20$ multivariate t-distributions with randomized means and covariance matrices and degree of freedom $2$ to create non-trivial densities $p(x|\theta)$. 
We note that the actual posterior density $p(\theta|x)$ (for any noise outputs $x_{i>8}$) retains the shape of the original SLCP model due to 
\begin{align}
    p(\theta|x) = \frac{p(x|\theta) p(\theta)}{p(x)} = \frac{p(x_{i>8}| x_{i\leq8}, \theta) p(x_{i\leq8}|\theta) p(\theta)}{p(x_{i>8}| x_{i\leq8}) p(x_{i\leq8})} = \frac{p(x_{i>8}) p(x_{i\leq8}|\theta) p(\theta)}{p(x_{i>8}) p(x_{i\leq8})} = \frac{ p(x_{i\leq8}|\theta) p(\theta)}{ p(x_{i\leq8})}
\end{align}

To avoid effects of the autoregressive nature of MAF density estimators, we re-order the dimensions $i$ of the original eight output dimensions $x_{i\leq8}$ and our added simulator outputs $x_{i>8}$ with a fixed random permutation.

Note that in Fig. \ref{fig:noisedims} the MMD for the true posterior (lower gray lines) is nonzero due to a finite number of samples being used.
\subsubsection{Lotka-Volterra model}
\label{supp:lv}

For the comparisons against previous neural conditional density models, we apply APT to infer the posterior of the Lotka-Volterra model as described in \cite{papamakarios18}.

\begin{SCfigure}[][h]
    \centering
    \includegraphics[width=3.25in]{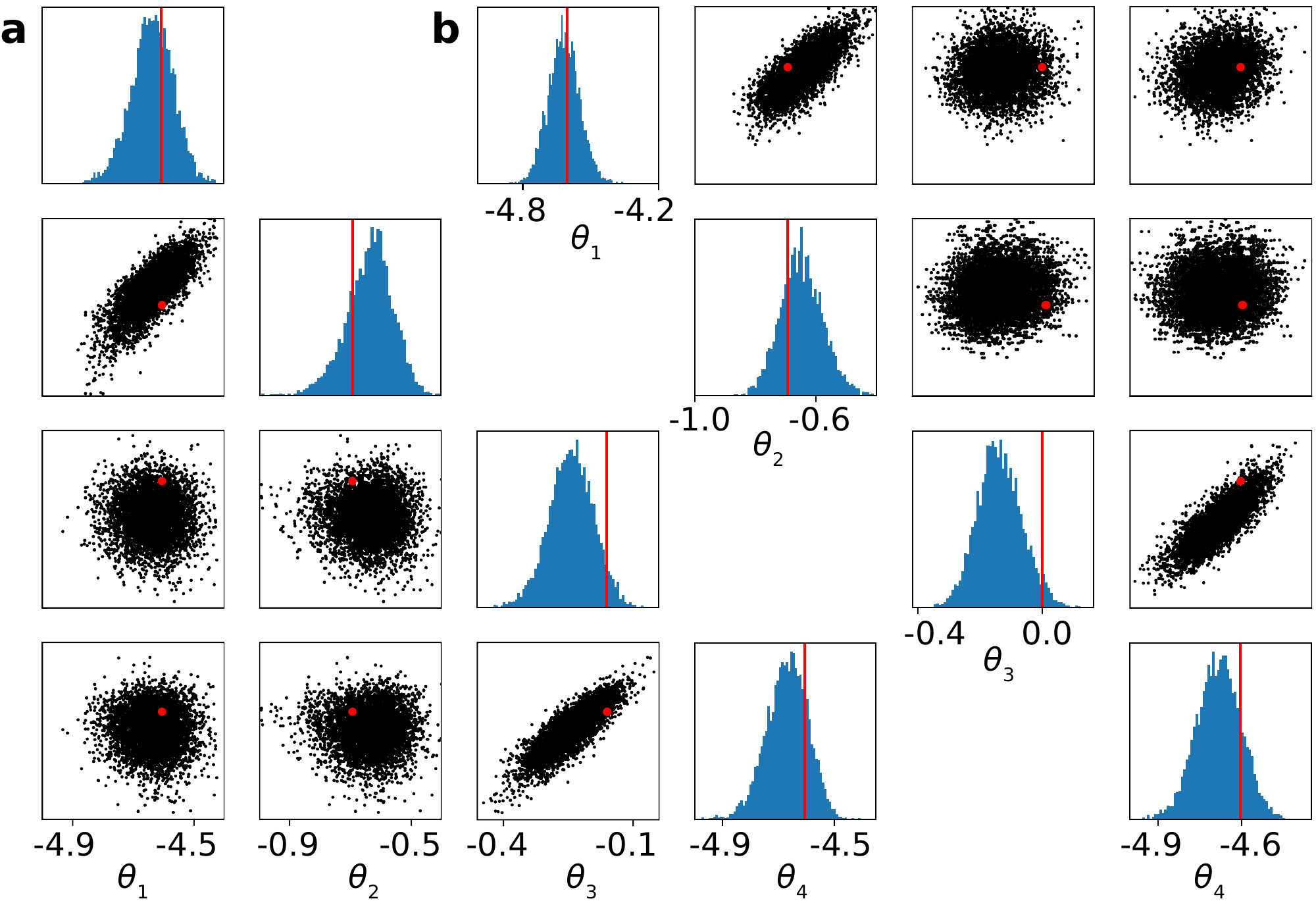}
    \label{fig:snl_exps_lv_lprobs}
    \caption{Lotka-Volterra model. Close-up comparison between \textbf{a} APT and \textbf{b} SNL posterior estimates.}
    \label{fig:lv_supp}
\end{SCfigure}
We also infer posteriors for a Lotka-Volterra model with added observation noise. We add independent Gaussian noise $\varepsilon_{it} \sim \mathcal{N}(0, \sigma^2)$ to both populations $i = 1,2$ and every time point $t=0, \ldots, 150$ of the raw simulated time-series. For RNN-APT, we added an initial layer of $100$ GRU units \cite{cho2014learning} to a MDN with a single Gaussian component.

\subsubsection{Rock-paper-scissors model}
\label{supp:rps}
We approximated the SPDE using a system of coupled stochastic differential equations as described in eq. (29-30) of \cite{reichenbach2008self}, and integrated this system using the Euler-Murayama method with a step size of $1$ on a 100x100 grid. We calculated second spatial derivatives using simple finite differences-of-differences. We initialized the system at the unstable uniform steady state, and integrated from $t=0$ to $t=100$. With $\mu$, $\sigma$ and $D$ denoting the growth rate, predation rate and diffusion constant as defined in \cite{reichenbach2007mobility, reichenbach2008self}, $\theta_1$, $\theta_2$ and $\theta_3$ were defined as their respective base 10 logarithms. We used uniform priors on each $\theta_j$, from $-1$ to $1$ for $1\leq j \leq 2$ and from $-6$ to $-5$ for $j=3$.

We used a CNN consisting of 6 convolutional layers with ReLu units, with each convolutional layer followed by a max pooling layer. The number of channels after each layer was: 8, 8, 8, 16, 32 and 32. The convolutional filter sizes were: 3, 3, 3, 3, 2 and 2. The max pooling sizes were: 1, 3, 2, 2, 2 and 1. The image sizes were 100x100, 32x32, 15x15, 6x6, 2x2 and 1x1. The 32-dimensional output of the CNN was then passed through two fully-connected tanh layers with with $50$ units each. We used a single-component MDN for this problem. Overall, for this architecture the elements of $\phi$ (weights and biases) numbered 8768 for the CNN layers, 4200 for the fully connected layers and 612 for the final mapping onto $\psi$.

\subsubsection{M/G/1 model}
\label{supp:MG1}
\begin{SCfigure}[][h]
	\centering
	\includegraphics[width=3.25in]{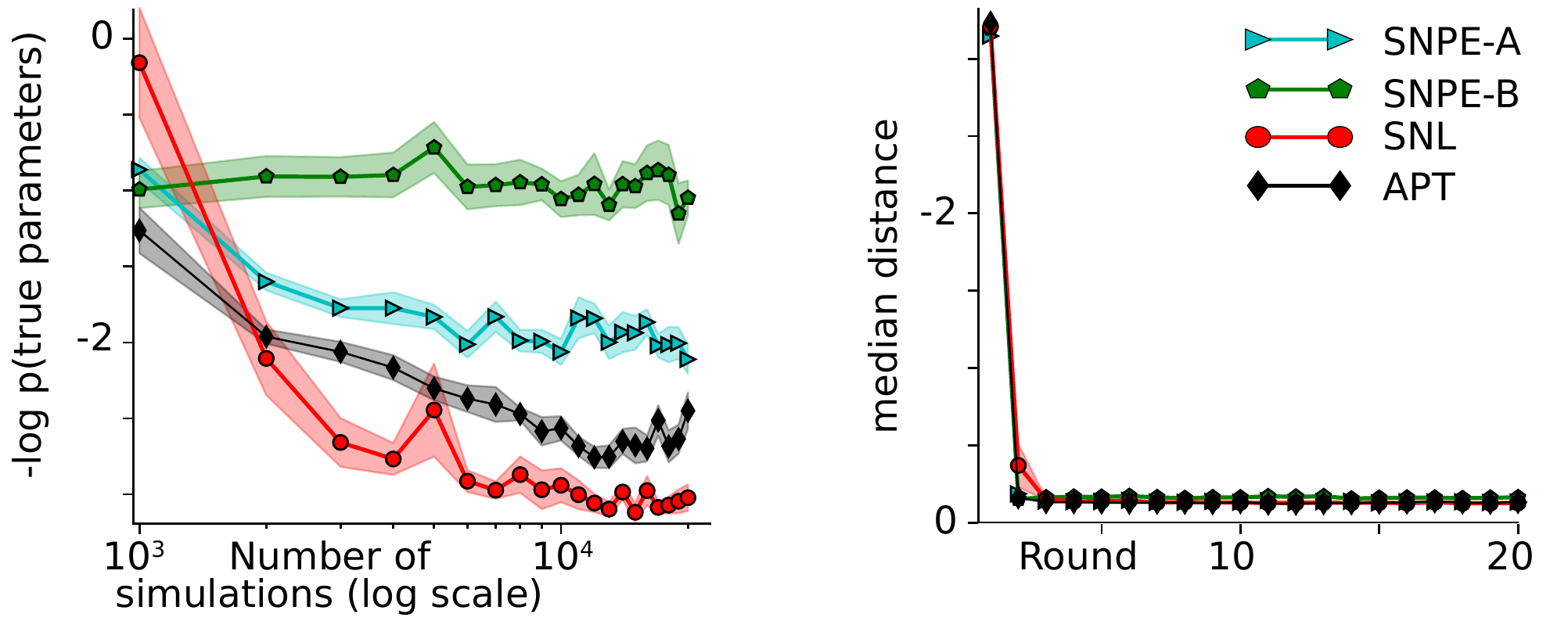}
    \label{fig:snl_exps_mg1_lprobs}
    \caption{M/G/1 model. Averages $\pm 1$ SEM over $10$ different random intializations with identical $x_o$ ($\leq10$ for SNPE-A in later rounds).}
    \label{fig:mg1_supp}
\end{SCfigure}

We compared MAF-APT to previous approaches on the M/G/1 queue model \citep[as described in][]{papamakarios16epsfree}. The results (mean $\pm$ SEM) for $10$ different random initializations with the identical $x_o$ are shown in figure \ref{fig:mg1_supp}. 

\subsubsection{Generalized linear model}
\label{supp:GLM}

We also compare MAF-APT against previous neural conditional density approaches on a Generalized Linear model model with a length-9 temporal input filter and a bias weight (i.e. $\theta_j, j=1,\ldots,d$ with $d=10$ parameters in total). We simulate $100$ time bins of activity in response to white noise and summarize the output with $10$ sufficient statistics $x$ as in \cite{lueckmann17}. We also train the algorithms with $5$ rounds of $N=5000$ each. We note that the posteriors for this simulator are well approximated as Gaussian, and use a single Gaussian component for the MDNs for SNPE-A and -B, and MAFs consisting of two MADES for SNL and APT. 
We use networks with two layers of $10$ tanh units for MDNs and MADES.

\begin{SCfigure}[][h]
    \centering
    \includegraphics[width=1.4in]{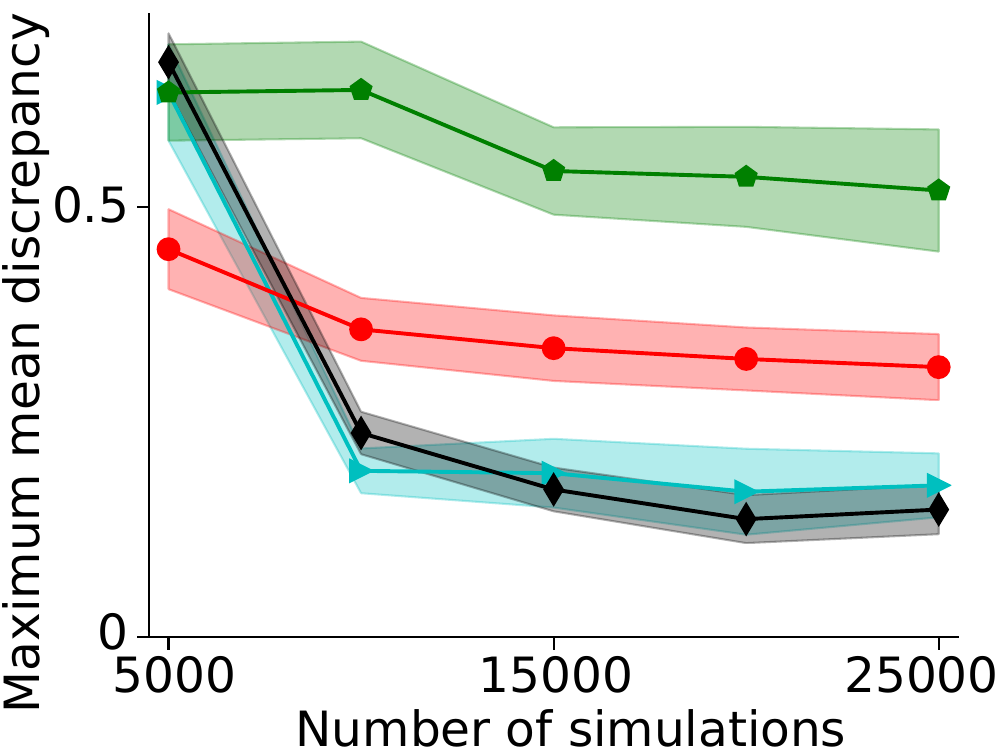}
    \label{fig:snl_exps_glm_mmds}
    \caption{Generalized linear model. Averages $\pm 1$ SEM over $10$ different random intializations.}
    \label{fig:glm_supp}
\end{SCfigure}

The results (mean $\pm$ SEM) for $10$ different random initializations and $x_o$ are shown in figure \ref{fig:glm_supp}. 

\end{document}